\newcommand{\model}{\texttt{MemOps}}

\documentclass[]{memtensor}
\usepackage[utf8]{inputenc}
\AtBeginDocument{%
  }

\usepackage{enumitem}
\usepackage{booktabs}
\usepackage{makecell}
\usepackage{CJKutf8}
\usepackage{booktabs}
\usepackage{pifont}
\usepackage[nointegrals]{wasysym}
\usepackage{amssymb}
\usepackage{graphicx}
\usepackage{xcolor}
\usepackage{colortbl}
\newcommand{\titlelogo}{%
  \raisebox{-0.15\height}{\includegraphics[height=1.05em]{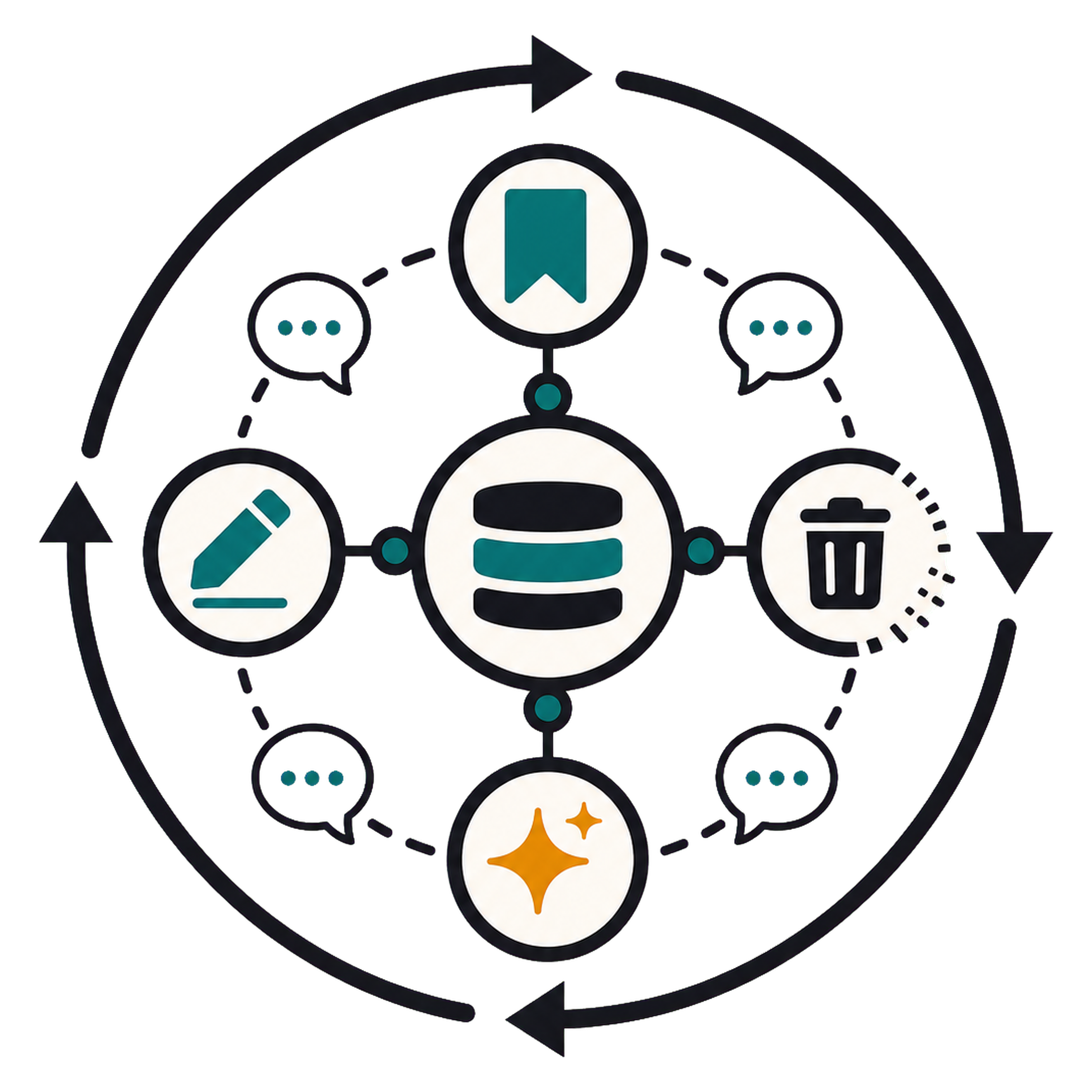}}%
  \hspace{0.35em}%
}

\title{\titlelogo MemOps: Benchmarking Lifecycle Memory Operations in Long-Horizon Conversations}
\author[1,2]{Xixuan Hao}
\author[1,3]{Zeyu Zhang}
\author[1]{Zehao Lin}
\author[1]{Yihang Sun}
\author[1]{Ziliang Guo}
\author[1]{Xichong Zhang}
\author[2\dag{}]{Yuxuan Liang}
\author[1]{Feiyu Xiong}
\author[1\dag{}]{Zhiyu Li}
\affiliation[1]{MemTensor (Shanghai) Technology Co., Ltd.}
\affiliation[2]{The Hong Kong University of Science and Technology (Guangzhou)}
\affiliation[3]{Renmin University of China}

\abstract{
Long-term memory has become a foundational capability for LLM-based agents that accompany users across extended, multi-session interactions.
Existing benchmarks, however, evaluate such memory almost exclusively through downstream question answering, scoring only the correctness of a final answer. 
This black-box formulation conflates the heterogeneous causes of memory failure, such as missing the introduction of a relevant fact, binding an operation to the wrong target, or relying on stale values after a correction.
As a result, it can credit correct answers despite their reliance on inconsistent or unsafe memory states.
In this paper, we argue that, in dynamic long-horizon interactions, memory is not a static collection of facts but a lifecycle of explicit operations, including remembering, forgetting, updating, reflecting, and their compositions. 
We introduce \model{}, a benchmark that reformulates conversational memory as a sequence of lifecycle operations and represents each memory event with a structured trace specifying its trigger, target, scope, state transition, and supporting evidence. 
A controllable generation pipeline embeds these operations into long, task-oriented conversations and produces gold operation traces together with six categories of operation-level probes, evaluated under both adjacent-evidence and long-context settings. 
Across long-context, retrieval-based, parametric and managed-memory systems, \model{} disentangles failure modes that final-answer accuracy alone conceals, revealing that current systems remain far from uniformly reliable. For instance, session-level retrieval outperforms turn-level retrieval, and long-context models remain notably weak at reconstructing ordered memory-state trajectories.
These results move long-term memory evaluation from final-answer scoring toward interpretable, operation-level diagnosis.
}

\date{\today}
\checkdata[Author Legend]{ $^\dagger$Corresponding authors (lizy@memtensor.cn,  yuxliang@outlook.com)}
\checkdata[Github]{\url{https://github.com/MemTensor/MemOps}}

\begin{document}
\maketitle


\section{Introduction}

In recent years, large language models (LLMs) have rapidly evolved from assistants confined to a single conversation session into interactive agents that accompany users across days, weeks, and even months of sustained interaction~\cite{wang2024survey,xi2025rise}. 
In this setting, long-term memory~\cite{zhang2025survey,sumers2024cognitive} has emerged as a foundational capability that determines how effectively LLM-based agents maintain coherent, personalized interaction over time.
Unlike assistants that respond solely to the current context, memory-augmented agents are expected to accumulate user-specific knowledge across interactions, retrieve relevant prior experiences, adapt to evolving user states, and disregard information that has grown obsolete or that the user has asked to be forgotten.
Motivated by this requirement, a growing body of benchmarks~\cite{locomo,wu2024longmemeval,tan2025membench} has substantially advanced the evaluation of long-term conversational memory, testing whether agents can recall facts, reason across sessions, handle temporal information, and answer questions grounded in long interaction histories.

\begin{figure}[t]
\centering
\includegraphics[width=0.95\columnwidth]{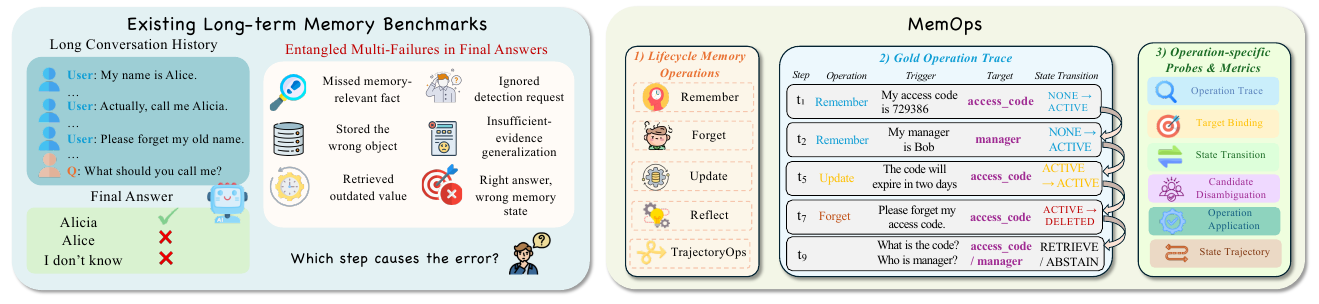}
\caption{Comparison between existing long-term memory benchmark and our \model.}
\vspace{-1em}
\label{fig:intro}
\end{figure}

Despite this progress, existing benchmarks~\cite{locomo,wu2024longmemeval,tan2025membench,ai2026memorybench,hu2026memoryagentbench,yang2026groupmembench,hu2026evermembench} largely evaluate long-term memory through downstream performance, most prominently question-answering (QA) tasks grounded in multi-session conversational histories.
As illustrated in Figure~\ref{fig:intro}, while this formulation yields an end-to-end measure of whether a system can exploit conversational histories to produce a correct response, final-answer accuracy alone often obscures the heterogeneous causes of memory failure.
An incorrect response may stem from errors at different stages of the memory process, such as missing the introduction of a memory-relevant fact, binding attributes to an incorrect object, or drawing unsupported generalizations from insufficient evidence.
Conversely, an agent may produce a correct answer while still operating over an incorrect or unsafe memory state, a discrepancy that remains difficult to diagnose when memory is evaluated solely as black-box input-output behavior.

This limitation is most pronounced in dynamic long-term interactions~\cite{sun2026preference}, where memory functions not as a static collection of facts but as a lifecycle process~\cite{lin2026memorysecurity}. Over the course of such interactions, a user may introduce a piece of information, subsequently correct it, request that part of it be forgotten, or implicitly signal a preference through recurring behavior. These events instantiate distinct memory operations, namely remembering, updating, forgetting, and reflecting, each defined by its own trigger, target object, scope, state transition, and characteristic failure modes.
For example, a successful forgetting operation requires not only withholding the forgotten value in future responses, but also preserving unrelated active memories that should remain available~\cite{maini2024tofu,shi2025muse}.
Similarly, a successful update requires distinguishing the current value from a stale historical value, rather than treating both as equally valid evidence~\cite{cohen2024ripple}. A reflective memory operation requires synthesizing multiple clues into a supported conclusion without over-inferring beyond the evidence. 
However, existing benchmarks evaluate these operations only through their downstream 
answers, offering no direct supervision or diagnosis of the operation itself.
As summarized in Table~\ref{tab:benchmark_comparison}, we compare \model{} against 
representative long-term memory benchmarks along five dimensions, including explicit 
memory operations, lifecycle coverage, state-transition modeling, forgetting and 
leakage control, and failure diagnosis.
While several benchmarks partially address 
individual dimensions, such as evaluating knowledge updates or selective forgetting 
as isolated capabilities, none treats the memory operation itself as the explicit 
unit of construction and supervision, nor jointly supports state-transition modeling 
and fine-grained failure diagnosis.

In this work, we propose \model, a lifecycle memory operations benchmark for diagnosing long-term memory behavior in memory-augmented agents. 
Rather than focusing solely on question answering over long conversations, \model~evaluates whether an agent can detect, execute, and respect explicit or implicit memory operations over conversational histories.
Our benchmark decomposes long-term memory capability into operation-level components and represents each memory event with a structured trace specifying its trigger, object, scope, state transition, and supporting evidence. This formulation makes it possible to evaluate not only the final answer but also the intermediate memory behavior that produced it.

To support this diagnostic objective, \model~constructs controlled long-term conversational histories in which memory operations are embedded in natural task-oriented interactions. Each example contains a gold memory operation trace and a set of metric-targeted probes. These probes evaluate operation detection, target extraction, state transition, operation-specific robustness, and provenance support. For instance, a forgetting example may first establish a temporary credential, introduce a separate retained control fact, and later request deletion of only the temporary credential. The evaluation then distinguishes leakage of the forgotten value from over-forgetting of the retained fact. Similarly, an update example distinguishes correct adoption of the new value from continued reliance on the stale value, while a reflection example tests whether the model can infer a preference from multiple clues without unsupported generalization.

Based on our benchmark, we evaluate representative systems from four canonical memory paradigms, including long-context models, retrieval-based memory, parametric memory, and managed memory, under both adjacent-evidence and long-context settings. The results show that performance is generally strongest when evidence is presented adjacent to the query, while dispersing that evidence into a larger, distractor-laden history consistently degrades both answer accuracy and operation-level reliability. Among the evaluated systems, session-level retrieval substantially outperforms turn-level retrieval, and managed-memory services that preserve longer, context-rich memory units outperform those that store short, isolated facts, indicating that surrounding context is critical for correctly executing lifecycle operations rather than merely retrieving relevant content. Parametric memory, which folds interaction history directly into model parameters, remains markedly unreliable across almost all diagnostic dimensions. Notably, operation-level analysis reveals that reconstructing an ordered memory-state trajectory across multiple composed operations is far more fragile than any single-step operation, and that this weakness persists even for otherwise strong models, exposing a failure mode that final-answer accuracy alone would not surface.

Our contributions can be summarized into three parts: 
\begin{itemize}[leftmargin=*]
    \item \textbf{A lifecycle operation formulation of conversational memory.} We introduce a lifecycle operation formulation of conversational memory, decomposing long-term memory into explicit operations with structured triggers, targets, scopes, state transitions, and failure types. 
    \item \textbf{A controllable benchmark generation pipeline}. 
    We develop a controllable benchmark generation pipeline that embeds memory operations into long-term task-oriented conversations and produces gold operation traces together with operation-specific probes. 
    \item \textbf{Evaluation and findings.} We show that performance degrades once evidence is diluted 
by long, distractor-laden histories. In addition, session-level retrieval and 
context-preserving memory outperform fine-grained storage. Finally, reconstructing 
ordered memory-state trajectories remains a persistent weakness even for strong 
models, revealing failures that final-answer accuracy alone cannot surface.
\end{itemize}

\newcommand{\cmark}{{\color[HTML]{2191A8}\ding{51}}}
\newcommand{\xmark}{{\color[HTML]{E7524C}\ding{55}}}
\newcommand{\pmark}{{\color[HTML]{2191A8}\LEFTcircle}}

\begin{table*}[t]
\centering
\small
\caption{
Comparison between \model~and existing long-term memory benchmarks.
\model~uniquely emphasizes explicit memory operations, lifecycle state transitions,
safe forgetting, leakage control, and fine-grained failure diagnosis.
\cmark~=~explicit support;~\pmark~=~partial support;~\xmark~=~not supported.
Ops denotes Operations.}
\resizebox{\linewidth}{!}{
\begin{tabular}{lccccc}
\toprule
\textbf{Benchmark} &
\textbf{Explicit Memory Ops} &
\textbf{Lifecycle Ops} &
\textbf{State Transition} &
\textbf{Forget / Leakage} &
\textbf{Failure Diagnosis} \\
\midrule
LoCoMo~\cite{locomo}              & \xmark & \xmark & \xmark & \xmark & \xmark \\
LongMemEval~\cite{wu2024longmemeval}        & \xmark & \pmark & \pmark & \pmark & \pmark \\
MemBench~\cite{tan2025membench}           & \xmark & \pmark & \pmark & \xmark & \pmark \\
MemoryBench~\cite{ai2026memorybench}        & \xmark & \pmark & \pmark & \pmark & \xmark \\
MemoryAgentBench~\cite{hu2026memoryagentbench}   & \xmark & \pmark & \cmark & \cmark & \pmark \\
GroupMemBench~\cite{yang2026groupmembench}      & \xmark & \pmark & \pmark & \pmark & \pmark \\
EverMemBench~\cite{hu2026evermembench}       & \xmark & \pmark & \pmark & \pmark & \pmark \\
\midrule
\textbf{MemOps} & \cmark & \cmark & \cmark & \cmark & \cmark \\
\bottomrule
\end{tabular}
}
\label{tab:benchmark_comparison}
\end{table*}

\section{Related Work}

\subsection{Long-Term Conversation Benchmarks}

Evaluation of conversational memory has progressively shifted from
short, single-session dialogue modeling toward question answering over
long, accumulated user-assistant histories. LoCoMo~\cite{locomo} marks
one end of this trend by generating very long conversations that span
hundreds of turns and tens of sessions, grounding them in fixed personas
and temporal event graphs, and probing memory through question
answering, event summarization, and multimodal generation. Its analysis
shows that long-context models and retrieval-augmented readers remain
well below human performance, particularly on temporal and adversarial
questions. 
LongMemEval~\cite{wu2024longmemeval} organizes evaluation around five core abilities: information extraction, multi-session reasoning, temporal reasoning, knowledge update, and abstention. It conceptualizes a memory system as a three-stage process comprising indexing, retrieval, and reading, and constructs length-configurable histories to assess these capabilities.
Both benchmarks establish the now-standard setting in which relevant evidence
is diluted by a long, noisy history, but they ultimately score a system
on the correctness of its final answer.

Another line of benchmarks expands the scope of memory evaluation beyond simple information retention.
MemBench~\cite{tan2025membench} distinguishes factual memory from
reflective memory and participation scenarios from observation
scenarios, and reports capacity and efficiency alongside accuracy.
MemoryBench~\cite{ai2026memorybench} reframes the task as continual learning from test-time user feedback. By separating declarative and procedural knowledge, it simulates both explicit and implicit feedback, evaluating whether a system can improve through accumulated interactions rather than relying solely on static retrieval.
MemoryAgentBench~\cite{hu2026memoryagentbench} frames evaluation as incremental multi-turn interaction and defines four core competencies, including accurate retrieval, test-time learning, long-range understanding, and selective forgetting. It argues that static long-context question answering fails to capture how a memory agent continuously ingests and updates information over time.

Recent benchmarks have further extended memory evaluation to multi-party and collaborative settings.
GroupMemBench~\cite{yang2026groupmembench} studies multi-party conversations, where an identical utterance can imply different memory updates depending on the speaker. It emphasizes speaker-grounded belief tracking and shows that a plain retrieval baseline can match learned memory systems, suggesting that current ingestion pipelines discard important cues for group memory.
EverMemBench~\cite{hu2026evermembench} targets long-horizon collaborative memory across multiple participants and groups, evaluating fine-grained recall, memory awareness, and profile understanding at the million-token scale.

Despite their growing scope, these benchmarks share a common evaluation paradigm in which a history is paired with a question and the final answer is scored.
Even categories whose names denote specific operations, such as knowledge update in LongMemEval, selective forgetting in MemoryAgentBench, and user knowledge update in EverMemBench, are still measured through downstream question answering.
As a result, an incorrect answer cannot be attributed
to a particular stage of the memory process, and a correct answer may
rest on an unsafe or inconsistent memory state. 
Instead, our benchmark reformulates conversational memory as a sequence of explicit lifecycle operations and evaluates whether systems can detect, execute, and apply these operations through structured traces and operation-specific probes.

\subsection{Memory-Enhanced LLM Systems}

Existing approaches for equipping LLMs with long-term memory can mainly be divided into four categories.
The most straightforward approach retains the entire interaction history within the context window and relies on a long-context model to attend over the accumulated information.
While this design eliminates the need for external memory mechanisms, its inference cost grows with the length of the interaction history.
In addition, their ability to effectively utilize distant evidence deteriorates as the context expands~\cite{bai-etal-2024-longbench}, leading to the \textit{lost-in-the-middle} effect~\cite{wu2024longmemeval,hu2026memoryagentbench,liu-etal-2024-lost,hsieh2024ruler}.
Crucially, raw context does not explicitly indicate whether earlier statements have been corrected or retracted, leaving superseded information equally accessible to the model.

Retrieval-based approaches store past interactions in an external memory~\cite{xu2025amem,rezazadeh2025memtree} and retrieve relevant fragments at query time~\cite{park2023generative}, typically with sparse or dense similarity search.
MemoryBank~\cite{zhong2024memorybank} is representative of this design, where the granularity of stored information, whether at the session, turn, or fact level, materially affects both recall and reading~\cite{wu2024longmemeval}.
Because retrieval is primarily driven by semantic similarity, it may overlook information validity and temporal relevance.
As a result, such systems may retrieve stale or semantically confusable content, introducing the very distractors that operation-level evaluation aims to identify.

A different line of work integrates memory directly into the model itself through dedicated memory modules~\cite{wang2023longmem,zhang2026nextmem} or parameter updates~\cite{lu2026locas,meng2022rome}. Self-updatable architectures such as MemoryLLM~\cite{wang2024memoryllm} exemplify this direction by absorbing new information over time.
These designs achieve more compact memory representations than storing the full interaction history, but their internal states are less transparent and less amenable to fine-grained control, limiting targeted operations such as forgetting a single value while preserving related information.
More recently, agentic systems have treated memory as an explicitly managed store~\cite{rasmussen2025zep} with read, write, update, and delete operations. Representative examples include MemGPT~\cite{packer2023memgpt}, Mem0~\cite{chhikara2025mem0}, and MemOS~\cite{li2025memos}, which provide explicit interfaces for managing memory contents through operations such as creation, retrieval, modification and deletion.

\section{MemOps}

\model{} evaluates long-term conversational memory as a sequence of explicit lifecycle operations, rather than treating it as a static QA task.
Given a long-horizon user-assistant interaction history, a memory-augmented agent is required to determine which user statements should update its memory state, identify the target to which each update applies, specify the value to be introduced or removed, and select the evidence that supports the final response.
This paradigm complements long-memory QA benchmarks~\cite{locomo,wu2024longmemeval,tan2025membench} by moving beyond final-answer correctness to diagnose which step in the memory lifecycle succeeds or fails.

\subsection{Problem Definition}

{
In our work, each benchmark instance is represented as a tuple $(b, C, O, A)$. Here, $b$ is a
topic-specific user background, $C$ is a set of evidence conversations, $O$ is a
gold operation trace, and $A$ is a set of evaluation probes with expected
answers.
The operation trace is the central evaluation anchor.
Each operation contains an operation type, a target object, an
old value, a new value, and evidence spans. 
Evidence spans are grounded in exact user dialogue turns, ensuring that every evaluated memory state can be traced back to concrete dialogue evidence rather than relying on background metadata or assistant-generated summaries.
}

Specifically, the \model{} benchmark evaluates five types of lifecycle memory operations:
\textbf{(1) Remember} captures clean first-time memory establishment by introducing a user-provided fact as active memory, and tests whether the system misses memory-relevant facts or binds them to the wrong target.
\textbf{(2) Forget} removes a previously active value while preserving unrelated retained memories, thereby testing both leakage of the forgotten value and over-removal of memories that should remain active.
\textbf{(3) Update} replaces an earlier active value with a corrected or superseding value, exposing failures in which the system continues to rely on the stale value after the correction.
\textbf{(4) Reflect} converts multiple observed clues into a bounded inferred memory, with representative failures including unsupported over-generalization or failure to recover the inference justified by the evidence.
\textbf{(5) TrajectoryOps} composes multiple remember, update, forget, and reflect events over one or more targets across time, allowing the benchmark to evaluate whether a system can recover not only the final active state, but also intermediate states and the temporal order of operations.

\subsection{Benchmark Construction}
\label{sec:benchmark-curation}

\begin{figure}[t]
\centering
\includegraphics[width=0.95\columnwidth]{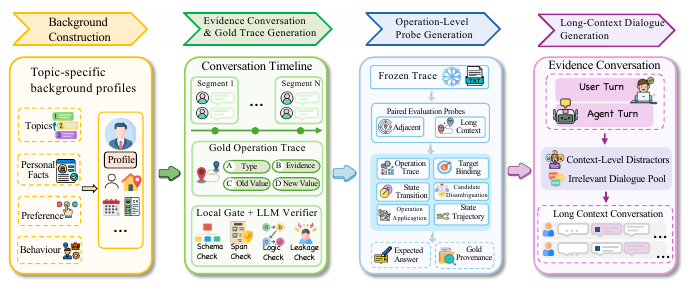}
\caption{Generation pipeline of our \model{} benchmark.}
\vspace{-1em}
\label{fig:memoryops-generation-pipeline}
\end{figure}

In this work, \model{} constructs each sample through a controlled generation and verification
pipeline, as shown in \Cref{fig:memoryops-generation-pipeline}. 
The pipeline has
four stages: \ding{182} Background Construction, \ding{183} Evidence Conversation and Gold Trace Generation, \ding{184} Operation-Level Probe
Generation, and \ding{185} Long-Context Dialogue Generation.
This staged design decouples scenario planning, evidence grounding, question construction, and long-horizon evaluation, ensuring that each retained example exhibits both natural conversational structure and explicit operation-level supervision. 
We describe the details of each stage as follows.

\textbf{Stage 1: Background Construction.} 
Our pipeline begins with topic-specific background profiles that describe user-centric attributes such as personal facts, life events, preferences, and potential behavioral cues. 
These backgrounds serve as the source material from which operation scenarios are sampled. For instance, a background may include a current address, a corrected ZIP code, a temporary permit that later becomes obsolete, commuting details, and preference signals related to quiet environments. 
The background functions solely as metadata to guide scenario generation and is not treated as evidence during evaluation.

\textbf{Stage 2: Evidence Conversation and Gold Trace Generation.}
For each background topic, we construct five corresponding dialogues, each incorporating one type of lifecycle memory operation.
For each topic-operation pair, \model{} generates an evidence conversation that
contains three segments. Each segment contains eight alternating turns,
starting with the user and ending with the assistant. User turns introduce
constraints, corrections, deletion requests, preferences, or supporting clues,
while assistant turns perform natural task-oriented work grounded in the user
messages. This structure makes the evidence compact enough for manual
inspection, while still requiring models to process information across multiple
conversation segments.

The evidence generator simultaneously produces the gold operation trace, where each operation includes a trigger span and supporting evidence, whose quoted text must appear verbatim in user turns. This constraint prevents the benchmark from relying on hidden labels or assistant paraphrases and enables fine-grained retrieval diagnosis. 
Specifically, in cases where a system produces an incorrect answer, it is possible to analyze whether the failure arises from retrieving the correct user turn, detecting the appropriate operation, binding the operation to the correct target, or applying the correct state transition.

\textbf{Stage 3: Operation-Level Probe Generation.}
After the evidence trace is finalized, a separate probe-generation stage creates
the evaluation questions and expected answers. 
We define six categories of operation-level probes to comprehensively assess different aspects of memory capability: 
\begin{itemize}[label=\ding{70},leftmargin=*]
    \item \textbf{OperationTrace} evaluates whether the model can identify which user messages trigger memory operations in the dialogue and determine what information within those messages should be recorded, updated, forgotten, or used for inference. The focus is not on producing the final answer, but on locating the relevant evidence and characterizing the corresponding operation.

\item \textbf{TargetBinding} evaluates whether the model can correctly bind each operation to its corresponding memory target. In a single dialogue, multiple entities such as addresses, individuals, preferences, or credentials may be present. The model must identify which target entity (and at what granularity) is affected by the operation, rather than conflating it with nearby or related entities.

\item \textbf{StateTransition} evaluates whether the model can infer the resulting memory state after an operation. For example, after a Remember operation, what should be stored; after an Update, whether the old value becomes invalid and what the new value is; after a Forget, whether the information remains accessible; and after a Reflect operation, which inferences are supported versus over-generalized. This probe focuses on reasoning about how the memory state evolves before and after each operation.

\item \textbf{CandidateDisambiguation} evaluates the model’s ability to disambiguate among multiple confusable candidates and reject distractors. Such distractors may include stale values, neighboring targets, assistant echoes, third-party information, or tentative values.

\item \textbf{OperationApplication} evaluates whether the model can apply the current memory state to downstream tasks such as form filling, reminder generation, drafting explanations, recommendation, computing reimbursements, or composing multiple facts. Compared to \textbf{StateTransition}, this probe goes beyond state identification and assesses whether the model can correctly utilize the state to complete real-world tasks.

\item \textbf{StateTrajectory} is used only for TrajectoryOps. It evaluates whether the model can track state evolution across multiple operations and recover both intermediate and final states. The focus is on reconstructing the full or partial state trajectory rather than inferring the state after a single operation.
\end{itemize}

In addition, each probe also includes gold provenance. The provenance is represented by exact user quotes and their segment-turn locations and is used for both answer generation and evaluation. 
When a probe requires multi-hop reasoning, the expected answer also includes an auditable reasoning chain that specifies the source facts, intermediate inferences, and the final derived answer~\cite{trivedi2022musique}. This design prevents nominally correct answers from being accepted when they are not supported by the required evidence path.

\textbf{Stage 4: Long-Context Dialogue Generation.}
After that, to simulate realistic long-context scenarios, we embed the evidence conversation into an unrelated conversation history. 
The pipeline selects a pool of irrelevant user-assistant dialogues and disperses the evidence segments across different conversations.
Each evidence segment is inserted at a role-compatible position so that the
resulting dialogue still alternates naturally between user and assistant turns.
The construction records where each evidence segment is inserted, including the
source conversation, segment index, insertion position, and message count.
In addition, to increase evaluation challenge, we construct context-level distractors, which are operation-aware noisy dialogue snippets inserted into the long-context history. Unlike irrelevant conversations, they are derived from the gold operation trace and remain semantically close to the target memory while being state-neutral.

The final benchmark supports two evaluation settings: adjacent setting and long-context setting. 
In the adjacent setting,
the model receives the evidence conversations directly before answering the
operation-level probes. 
This setting isolates the model’s ability to understand and apply memory operations when the relevant evidence is adjacent to the query.
In the long-context setting, the evidence segments are dispersed into a larger pool of unrelated user-assistant conversations. 
The resulting history records which irrelevant conversations received evidence, where the evidence was inserted, and which evidence segment was inserted. This setting tests whether memory systems can retain, retrieve, and apply operation evidence despite long-horizon noise.

\begin{figure}[t]
\centering
\includegraphics[width=0.95\columnwidth]{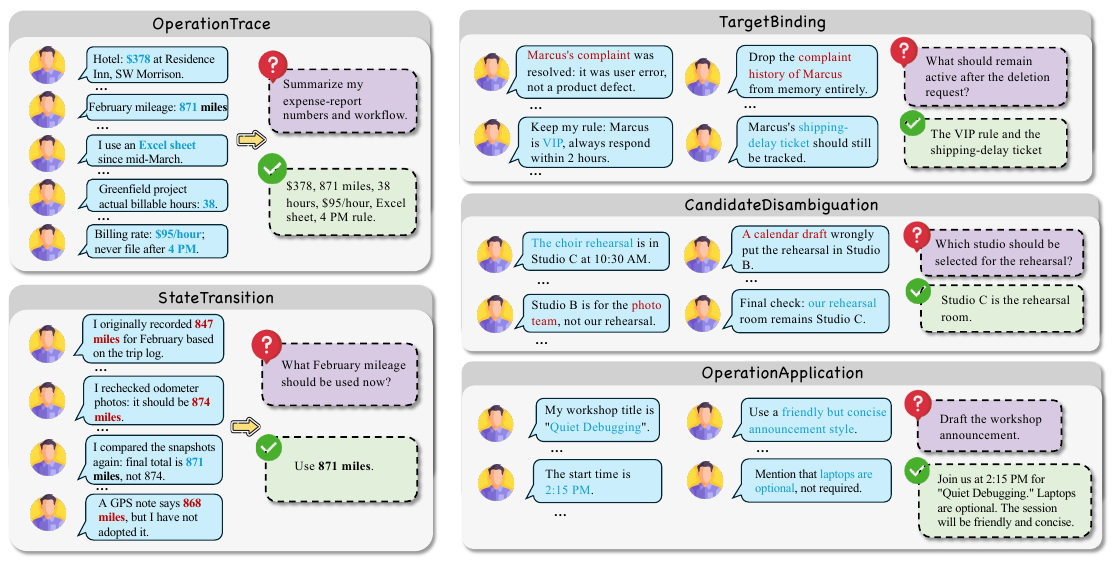}
\caption{Illustrative conversation examples for each operation-level probe type.}
\label{fig:memoryops-conversation-example}
\end{figure}
\subsection{Quality Verification}

To guarantee both semantic correctness and structural validity, \model{} combines deterministic local gates with LLM-based verification. The local gates first enforce a set of strict structural constraints, including schema validity, evidence formatting consistency, and operation-aware question structure. These checks reject malformed or schema-inconsistent samples prior to semantic verification.
The LLM verifier then audits higher-level semantic properties, including dialogue grounding, operation trace integrity, state transition consistency, and coverage of probe-relevant information, while also checking for potential leakage of internal labels.
If a sample fails either deterministic or LLM-based checks, the pipeline stores the failed attempt, converts the failure into feedback signals, and triggers regeneration. This design yields a benchmark in which each example exhibits both natural conversational surface form and machine-checkable operation semantics.

\subsection{Dataset Statistics}

Table~\ref{tab:dataset-overview} summarizes the overall scale, length, grounding, and quality-control statistics of \model{}. The benchmark spans 100 unique topics and comprises 403 evidence conversations, which decompose into 1,209 evidence-conversation segments and 9,672 dialogue turns. From these conversations we derive 2,006 unique QA pairs; since every pair is probed under both the adjacent and long-context settings, this yields 4,012 evaluation instances in total. Together with five lifecycle operation types and six operation-level probe types, and an average of 4.0 operations per trajectory, the benchmark covers a broad range of memory behaviors while keeping every trace compact enough for manual inspection.

\begin{figure}[t]
\centering
\includegraphics[width=0.95\columnwidth]{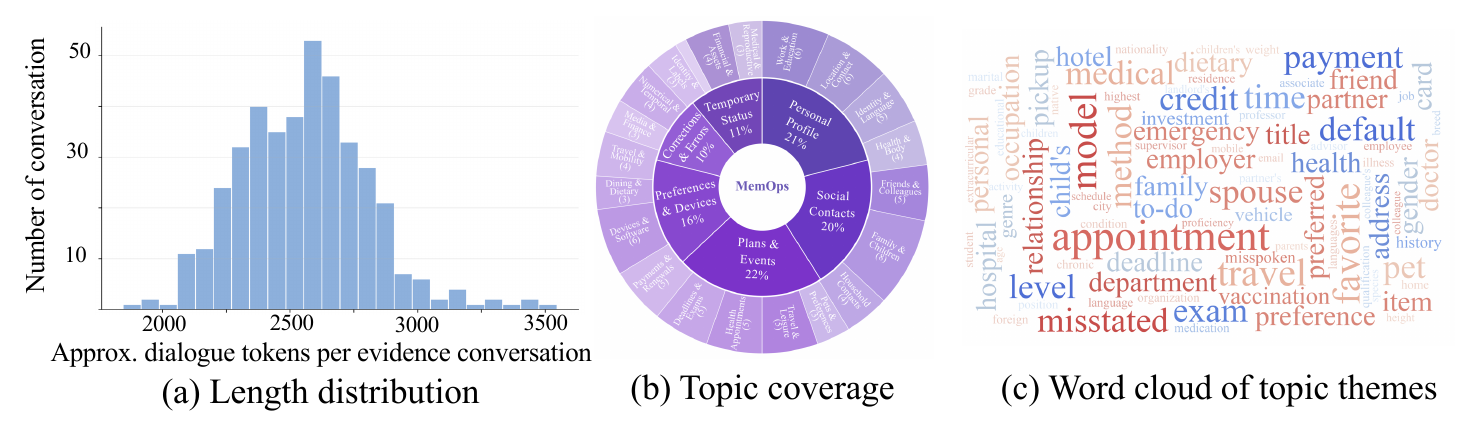}
\caption{Scale and topic characteristics of the \model{} dataset.}
\label{fig:question-type-distribution2}
\end{figure}

\textbf{Topic Coverage.}
As illustrated by the word cloud and the category breakdown in Figure~\ref{fig:question-type-distribution2}, the 100 topics are distributed across everyday memory-relevant domains rather than a single narrow scenario. The largest categories are Plans \& Events (22\%), 
Personal Profile (21\%) and Social Contacts (20\%), followed by Preferences \& Devices (16\%) , Temporary Status (11\%), and Corrections \& Errors (10\%), each further divided into fine-grained sub-topics such as health, addresses, and payment details. This spread ensures that lifecycle operations are instantiated over heterogeneous entities and attributes rather than a homogeneous fact type.

\begin{figure}[t]
\centering
\includegraphics[width=0.95\columnwidth]{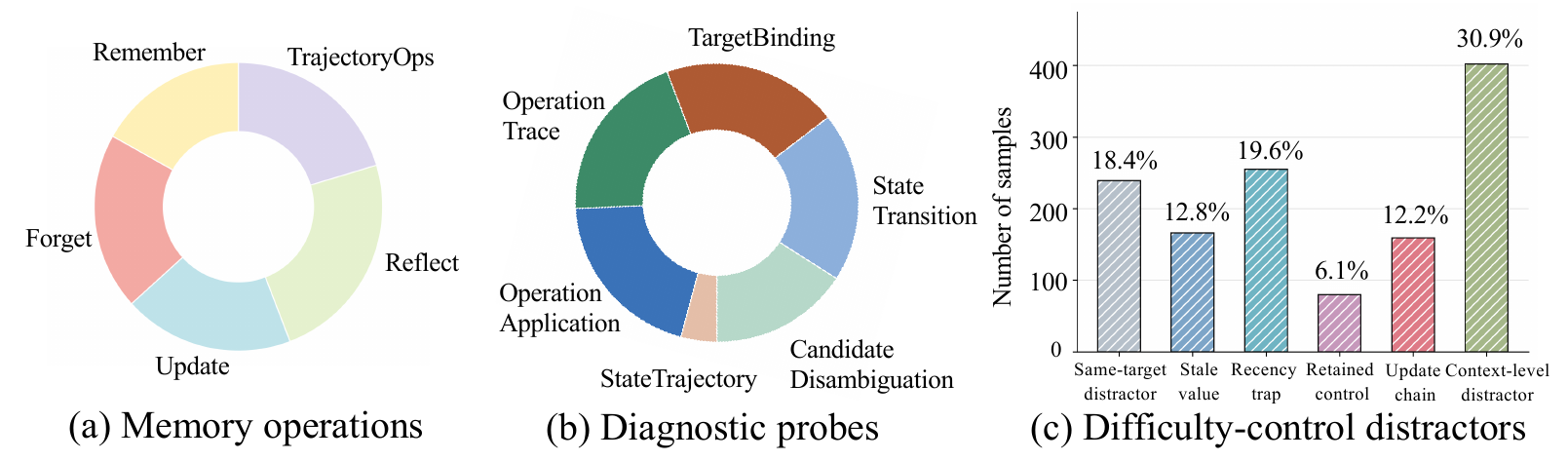}
\caption{Composition of \model{} operations and evaluation design.}
\label{fig:question-type-distribution}
\end{figure}

\textbf{Operation and Probe Distribution.}
We further report the composition of memory operations, evaluation probes, and targeted distractors. Figure~\ref{fig:question-type-distribution}(a) shows the five lifecycle operation types, namely remember, forget, update, reflect, and trajectoryops, which appear in comparable proportions, so that no single operation dominates the aggregate scores. Orthogonally, Figure~\ref{fig:question-type-distribution}(b) shows the six operation-level probe categories, namely OperationTrace, TargetBinding, StateTransition, CandidateDisambiguation, OperationApplication, and StateTrajectory, which are likewise broadly balanced, ensuring that each aspect of memory capability contributes meaningfully to the evaluation. Finally, Figure~\ref{fig:question-type-distribution}(c) reports how many samples are equipped with each type of targeted distractor or control condition, namely same-target distractor, stale value, recency-top, retained control, update chain, and context-level distractor. Context-level distractors are the most frequent, followed by recency-top and same-target distractors, so that every probe is stressed by evidence that is semantically close, temporally salient, or bound to a neighboring target rather than by shallow matching.

\textbf{Length Distribution.}
Figure~\ref{fig:question-type-distribution2} also shows the distribution of approximate dialogue tokens per evidence conversation. Lengths concentrate between roughly 2,000 and 3,500 tokens with a mean of 2,557, confirming that evidence traces are compact yet non-trivial. In the long-context setting, these evidence segments are dispersed within a substantially larger pool of unrelated conversations, raising the average context length to 60,821 tokens and up to 68,309 tokens at maximum, so the effective context presented to the model far exceeds the per-conversation figures reported above.

\textbf{Grounding and Difficulty.}
Beyond scale, \model{} is designed so that each probe requires integrating evidence rather than matching a single surface fact. On average, each probe is grounded in 3.47 evidence spans, and 20.1\% of probes demand multi-hop reasoning, chaining together several source facts to derive an intermediate inference. To further stress operation-level robustness, we insert an average of 5.7 context-level distractors per sample, snippets that are semantically close to the target memory yet state-neutral, ensuring that correct answers cannot be reached through shallow retrieval alone.

\begin{table}[t]
\centering
\small
\caption{Dataset statistics of \model{}.}
\label{tab:dataset-overview}
\begin{tabular}{l r p{2.5em} l r}
\rowcolor{black!10}
\toprule
\textbf{Statistic} & \textbf{Value} & & \textbf{Statistic} & \textbf{Value} \\
\midrule
\multicolumn{2}{l}{\textit{Scale}} & & \multicolumn{2}{l}{\textit{Length}} \\
Unique topics                       & 100     & & Avg.\ approx.\ tokens (adjacent)        & 2{,}557 \\
Evidence conversations              & 403     & & Avg.\ approx.\ tokens (long-context)    & 60{,}821 \\
Evidence-conversation segments      & 1{,}209 & & Max.\ approx.\ tokens (long-context)    & 68{,}309 \\
Dialogue turns                      & 9{,}672 & & \multicolumn{2}{l}{\textit{Grounding and difficulty}} \\
Unique QA pairs                     & 2{,}006 & & Avg.\ evidence spans per probe & 3.47 \\
Evaluation instances (adj.\ + long) & 4{,}012 & & Multi-hop probes (\%)          & 20.1 \\
\multicolumn{2}{l}{\textit{Operations and probes}} & & Avg.\ context-level distractors & 5.7 \\
Operation types                     & 5       & & \multicolumn{2}{l}{\textit{Quality control}} \\
Probe types                         & 6       & & Generated $\rightarrow$ retained & $727 \rightarrow 403$ \\
Avg.\ operations per trajectory     & 4.0     & & Retention rate (\%)            & 55.4 \\
\bottomrule
\end{tabular}
\end{table}

\section{Experiments}

\subsection{Experimental Setup}

Our goal is not to rank systems by a single leaderboard score, but to evaluate whether
\model{} can surface and localize failure modes throughout the memory lifecycle.
For each system, we therefore report both final-answer accuracy and a suite of
operation-level diagnostic metrics derived from the gold operation trace. This distinguishes systems that reach a correct answer
through a correct memory process from those that answer correctly while operating over
an inconsistent or unsafe memory state.

\textbf{Methods.}
We evaluate representative methods of the four memory families.
\textbf{(1)~Long-context memory}, which places the
entire interaction history in the context window and answers by direct attention. We
evaluate GPT-4o~\cite{hurst2024gpt4o}, GPT-4.1-mini~\cite{openaiIntroducingGPT41}, GLM-4.6~\cite{glm46}, Gemini-3-Flash~\cite{googleGemini3Flash}, Qwen3.6-27B~\cite{githubGitHubQwenLMQwen36}, DeepSeek-V4-Flash~\cite{xu2026deepseekv4}, and
Claude-Sonnet-4.5~\cite{anthropicClaudeSonnet}. \textbf{(2)~Retrieval-based memory}, which stores past interactions
externally and retrieves relevant fragments at query time; we use a BM25 retriever
backed by GPT-4.1-mini at two storage granularities, turn-level and session-level.
\textbf{(3) Parametric memory}, which folds interaction history directly
into model parameters; we test the parameter-update baseline Temp-LoRA~\cite{templora}. \textbf{(4) Managed memory}, which
maintains memory as an explicitly managed store with read, write, update, and delete operations; we test
the managed-memory services Mem0~\cite{chhikara2025mem0} and MemOS~\cite{li2025memos}.
Each long-context model is run under both evaluation
settings described below, while the retrieval and memory-system baselines follow their
native ingestion and query interfaces.

\textbf{Settings and Probes.}
Following the benchmark design in Section~\ref{sec:benchmark-curation}, each sample is evaluated
under two settings. In the \emph{adjacent} setting, the evidence conversation is
presented directly before the probe, isolating whether a system can detect and apply the
correct memory operation when the relevant evidence is local. In the \emph{long-context}
setting, the evidence segments are dispersed within a large pool of unrelated
user--assistant conversations, which we draw from UltraChat~\citep{ultrachat}, so that
relevant evidence is diluted by realistic long-horizon noise. Across 403 evidence conversations, \model{} contains 2,006 unique question–answer pairs with operation-specific probe compositions. Each pair is instantiated under both settings, yielding 4,012 evaluation instances.
The paired design measures the same underlying capability
both immediately after the evidence and after that evidence has been embedded in the long
noisy history, so that any degradation can be attributed to long-context interference
rather than to the underlying operation.



\textbf{Metrics.}
We report answer \textit{accuracy} as the primary task-level metric. Since \model{} is designed to evaluate memory operations rather than focusing solely on final outputs, we additionally report a set of diagnostic metrics derived from the gold operation trace.
\underline{Operation F1} measures whether the model correctly identifies memory-relevant events. 
\underline{Provenance support} assesses whether the generated answer is grounded in the required user-provided evidence. 
\underline{Leakage Rate} quantifies the extent to which forgotten values are inappropriately reused, while \underline{Stale Value Rate} measures cases where updated values are incorrectly replaced with earlier, obsolete ones.
\underline{Reflect Precision} assesses whether inferred reflection memories are evidence-supported, correctly scoped, and not overgeneralized beyond the relevant memory target.

For these operation-level diagnostics, an LLM judge based on GPT-4o~\cite{hurst2024gpt4o} compares each model response against the gold operation trace, gold provenance, and evidence conversation, assigns binary labels for the applicable metrics, and we report their averages over the evaluated examples.
Through this evaluation process, we turn long-term memory evaluation from a
single final-answer score into an auditable account of how a memory state was formed, changed, and used.

\subsection{Overall Results}
\begin{table*}[h!]
\centering
\small
\setlength{\tabcolsep}{4pt}
\caption{Overall performance on the \model{}. 
}
\label{tab:overall}
\resizebox{\textwidth}{!}{
\begin{tabular}{lcccccc}
\toprule
\rowcolor{black!10}
\textbf{Method} & \textbf{Accuracy} & \textbf{Operation F1} & \textbf{Provenance} & \textbf{Leakage} & \textbf{Stale Value} & \textbf{Reflect Precision} \\
\midrule
GPT-4o (Adjacent) & 0.850 & 0.798 & 0.884 & 0.132 & 0.045 & 0.729 \\
GPT-4.1-mini (Adjacent) & 0.879 & \textbf{0.888} & 0.926 & 0.265 & 0.052 & 0.820 \\
Claude-Sonnet-4.5 (Adjacent) & \textbf{0.916} & 0.887 & 0.941 & 0.230 & 0.045 & \textbf{0.874} \\
Gemini-3-Flash (Adjacent) & 0.848 & 0.798 & 0.895 & 0.113 & 0.033 & 0.778 \\
GLM-4.6 (Adjacent) & 0.888 & 0.816 & 0.911 & 0.192 & 0.045 & 0.847 \\
Qwen3.6-27b (Adjacent) & 0.914 & 0.875 & \textbf{0.944} & \textbf{0.095} & 0.029 & 0.864 \\
DeepSeek-V4-Flash (Adjacent) & 0.790 & 0.821 & 0.818 & 0.228 & 0.075 & 0.738 \\
\midrule
GPT-4o (Long Context) & 0.794 & 0.815 & 0.823 & 0.128 & \textbf{0.016} & 0.756 \\
GPT-4.1-mini (Long Context) & 0.798 & 0.825 & 0.830 & 0.250 & 0.029 & 0.814 \\
Claude-Sonnet-4.5 (Long Context) & 0.790 & 0.650 & 0.806 & 0.268 & 0.030 & 0.731 \\
Gemini-3-Flash (Long Context) & 0.663 & 0.673 & 0.746 & 0.116 & 0.030 & 0.630 \\
GLM-4.6 (Long Context) & 0.808 & 0.772 & 0.792 & 0.210 & 0.039 & 0.816 \\
Qwen3.6-27b (Long Context) & 0.790 & 0.698 & 0.584 & 0.128 & 0.026 & 0.730 \\
DeepSeek-V4-Flash (Long Context) & 0.798 & 0.716 & 0.804 & 0.163 & 0.023 & 0.777 \\
\midrule
RAG (GPT-4.1-mini) (Turn-Level) & 0.618 & 0.629 & 0.694 & 0.195 & 0.136 & 0.613 \\
RAG (GPT-4.1-mini) (Session-Level) & 0.845 & 0.845 & 0.859 & 0.247 & 0.042 & 0.859 \\
Mem0 & 0.543 & 0.534 & 0.556 & 0.135 & 0.102 & 0.559 \\
MemOS & 0.785 & 0.813 & 0.811 & 0.220 & 0.052 & 0.764 \\
Temp-LoRA & 0.162 & 0.559 & 0.182 & 0.165 & 0.238 & 0.146 \\
\bottomrule
\end{tabular}
}
\end{table*}

We demonstrate the overall performance across different models and evaluation settings in Table~\ref{tab:overall}. 
As we can see, the strongest performance is generally observed under the adjacent-evidence setting, because long-context settings introduce distractor and irrelevant contextual information, which increases reasoning interference and makes it harder to accurately identify memory operations and ground responses in relevant evidence. 
Claude-Sonnet-4.5 achieves the highest accuracy and the best reflect precision among all evaluated systems, narrowly ahead of Qwen3.6-27B. Qwen3.6-27B nonetheless stands out on the safety-oriented diagnostics, achieving the strongest provenance support and the lowest leakage rate, indicating a markedly more consistent and reliable memory-operation profile even though it trails Claude-Sonnet-4.5 on the top-line accuracy and reflect precision.
GLM-4.6 attains the third-highest accuracy, but its operation-level metrics remain comparatively modest. 
GPT-4.1-mini achieves the strongest Operation F1, yet its leakage rate and stale value rate remain less satisfactory. GPT-4o also remains competitive across metrics, with relatively balanced but not particularly outstanding performance. 
Overall, these results suggest that different models exhibit distinct strengths and behavioral patterns across evaluation dimensions, with Qwen3.6-27B standing out for combining high answer accuracy with consistently strong operation-level reliability.

Long-context evaluation generally yields lower accuracy than the adjacent-evidence setting, though the degree of degradation varies across models. 
Among long-context methods, GLM-4.6 achieves the highest answer accuracy, followed by DeepSeek-V4-Flash and GPT-4.1-mini. For DeepSeek-V4-Flash, we observe that the long-context setting slightly outperforms the adjacent-evidence setting. We attribute this pattern mainly to two factors. First, the LLM-based judge exhibits some instability, introducing noise into the evaluation results. Second, in the adjacent-evidence setting, outdated values, updated values, tentative values, and forgetting requests often appear in close proximity within the evidence chain. 
This encourages models to reproduce the evidence in greater detail, which can cause values that should have been suppressed to leak into the final answer or predicted operations. 
In contrast, the longer context encourages more compressed outputs that focus on the final state, reducing historical repetition and thereby lowering the observed leakage rate.

Retrieval-augmented and memory-system baselines exhibit substantial variation. Session-level RAG with GPT-4.1-mini substantially outperforms turn-level RAG, improving both answer accuracy and operation F1. 
This is because turn-level retrieval fragmenting evidence segments and losing the cross-turn context that operations like Update and Reflect require.
MemOS performs competitively across most metrics, whereas Mem0 is weaker. 
Specifically, MemOS returns long memories that stay close to the original dialogue, preserving the operation chain and surrounding context. Mem0, by contrast, returns shorter extracted memory representations; although retrieval does hit the correct session, it frequently loses the fine-grained details, ordering, and neighboring information that \model{} judging depends on.
Temp-LoRA performs poorly across almost all metrics. This is because the current Temp-LoRA implementation essentially encodes text transiently into the model’s parameter space, functioning as an implicit weight-level memorization mechanism rather than a reliable, controllable memory operation executor. 
While it offers marginal gains on shallow tasks such as short-fact recall and candidate selection, it remains poorly aligned with \model{} tasks that require state updates, evidence-boundary reasoning, trajectory-level inference, and application generation.

\subsection{Results by Operation Types}
\label{sec:operation_type}

\begin{table*}[h!]
\centering
\setlength{\tabcolsep}{10pt}
\caption{Answer accuracy by operation types.}
\label{tab:memoryops-v13-operation-type}
\footnotesize
\begin{tabular}{lccccc}
\toprule
\rowcolor{black!10}
\textbf{Method} & \textbf{Remember} & \textbf{Forget} & \textbf{Update} & \textbf{Reflect} & \textbf{TrajectoryOps} \\
\midrule
GPT-4o (Adjacent) & 0.902 & 0.898 & 0.877 & 0.746 & 0.856 \\
GPT-4.1-mini (Adjacent) & 0.934 & 0.872 & 0.886 & 0.823 & 0.890 \\
Claude-Sonnet-4.5 (Adjacent) & \textbf{0.946} & 0.875 & 0.916 & 0.898 & \textbf{0.949} \\
Gemini-3-Flash (Adjacent) & 0.882 & 0.820 & 0.880 & 0.796 & 0.880 \\
GLM-4.6 (Adjacent) &0.926&0.855&0.890&0.850&0.924 \\
Qwen3.6-27b (Adjacent) & 0.936 & \textbf{0.913} & \textbf{0.929} & 0.869 & 0.934 \\
DeepSeek-V4-Flash (Adjacent) & 0.855 & 0.768 & 0.776 & 0.748 & 0.805 \\
\midrule
GPT-4o (Long Context) & 0.868 & 0.815 & 0.802 & 0.756 & 0.737 \\
GPT-4.1-mini (Long Context) & 0.838 & 0.785 & 0.825 & 0.817 & 0.727 \\
Claude-Sonnet-4.5 (Long Context) & 0.811 & 0.735 & 0.838 & 0.752 & 0.829 \\
Gemini-3-Flash (Long Context) & 0.703 & 0.668 & 0.734 & 0.590 & 0.651 \\
GLM-4.6 (Long Context) & 0.836 & 0.780 & 0.841 & 0.821 & 0.768 \\
Qwen3.6-27b (Long Context) & 0.826 & 0.768 & 0.847 & 0.758 & 0.768 \\
DeepSeek-V4-Flash (Long Context) & 0.843 & 0.820 & 0.818 & 0.790 & 0.727 \\
\midrule
RAG (GPT-4.1-mini) (Turn-level) & 0.757 & 0.650 & 0.519 & 0.683 & 0.444 \\
RAG (GPT-4.1-mini) (Session-level) & 0.897 & 0.807 & 0.821 & \textbf{0.900} & 0.785\\
Mem0 & 0.645 & 0.580 & 0.510 & 0.550 & 0.424 \\
MemOS & 0.860 & 0.770 & 0.802 & 0.769 & 0.729 \\
Temp-LoRA & 0.191 & 0.103 & 0.250 & 0.123 & 0.173 \\
\bottomrule
\end{tabular}
\end{table*}
Table~\ref{tab:memoryops-v13-operation-type} reveals that accuracy varied substantially across operation types.
In the adjacent setting, Claude-Sonnet-4.5 and Qwen3.6-27B split dominance across the five operation types: Claude-Sonnet-4.5 led on Remember, Reflect, and TrajectoryOps, while Qwen3.6-27B led on Forget and Update, with GPT-4.1-mini, GLM-4.6, and GPT-4o trailing closely behind on individual operations.
This indicates that when evidence is presented locally, the two strongest models specialize somewhat, with Claude-Sonnet-4.5 holding a particular edge on operations that require synthesizing or ordering evidence over time in Reflect and TrajectoryOps.
In addition, Qwen3.6-27B is more reliable on operations that require precise removal or replacement of a single value in Forget and Update.
In contrast, performance in the long-context setting exhibited far greater variability across operation types, with no single model maintaining dominance.
GPT-4o led on Remember, DeepSeek-V4-Flash led on Forget, Qwen3.6-27B led on Update, GLM-4.6 led on Reflect, and Claude-Sonnet-4.5 led on TrajectoryOps.

We also find that context expansion erodes multi-step state-tracking ability.
As shown in Table~\ref{tab:memoryops-v13-operation-type}, TrajectoryOps consistently emerges as the most sensitive dimension, whether the comparison is made by moving the same model from the adjacent setting to the long-context setting, or by changing the same RAG pipeline from turn-level to session-level retrieval.
This pattern reflects the intrinsic sensitivity of these tasks to evidence organization.
Single-step operations such as Remember and Forget typically require locating a single evidence span and applying one state transition, making them relatively robust to interference.
By contrast, multi-step operations require the model to recover dispersed operations from a long, distractor-heavy context, order them temporally, and reconstruct the resulting memory-state trajectory.
This \underline{longer span, greater error accumulation} property makes trajectory-level reasoning especially vulnerable to both context dilution in long-context evaluation and evidence fragmentation under turn-level retrieval.

Moreover, we observe an anomalous pattern in which DeepSeek-V4-Flash achieves higher accuracy in the long-context setting than in the adjacent setting on Forget, Update, and Reflect operations.
We attribute this pattern to its distinctive long-context generation behavior. Unlike GPT-4o, GPT-4.1-mini, and GLM-4.6, DeepSeek-V4-Flash produces substantially more completion tokens in the long-context setting, while the average length of its final answers decreases~\cite{rodionov2026reasoning}. This suggests that, under long-context evaluation, DeepSeek allocates more generation budget to predicted operations, provenance identification, and state organization, rather than merely producing longer final answers. As a result, it may be better able to consolidate the currently valid memory state for operations such as Forget, Update, and Reflect.

\subsection{Results by Probe Types}

\begin{table*}[t]
\centering
\small
\setlength{\tabcolsep}{4pt}
\caption{Answer accuracy by probe types.}
\label{tab:memoryops-v13-eval-type}
\resizebox{\textwidth}{!}{
\begin{tabular}{lcccccc}
\toprule
\rowcolor{black!10}
\textbf{Method} & \makecell{\textbf{Operation}\\\textbf{Trace}} & \makecell{\textbf{Target}\\\textbf{Binding}} & \makecell{\textbf{State}\\\textbf{Transition}} & \makecell{\textbf{Candidate}\\\textbf{Disambiguation}} & \makecell{\textbf{Operation}\\\textbf{Application}} & \makecell{\textbf{State}\\\textbf{Trajectory}} \\
\midrule
GPT-4o (Adjacent) & 0.784 & 0.873 & 0.850 & 0.981 & 0.787 & 0.866 \\
GPT-4.1-mini (Adjacent) & 0.871 & 0.891 & 0.898 & 0.956 & 0.797 & 0.866 \\
Claude-Sonnet-4.5 (Adjacent) & 0.864 & \textbf{0.940} & 0.911 & \textbf{0.988} & 0.891 & \textbf{0.927} \\
Gemini-3-Flash (Adjacent) & 0.757 & 0.861 & 0.871 & 0.956 & 0.829 & 0.805 \\
GLM-4.6 (Adjacent) & 0.851 & 0.908 & 0.901 & 0.966 & 0.824 & 0.915 \\
Qwen3.6-27b (Adjacent) & \textbf{0.883} & 0.928 & \textbf{0.931} & 0.963 & 0.873 & 0.915 \\
DeepSeek-V4-Flash (Adjacent) & 0.670 & 0.856 & 0.815 & 0.938 & 0.710 & 0.744 \\
\midrule
GPT-4o (Long Context) & 0.603 & 0.873 & 0.858 & 0.984 & 0.772 & 0.390 \\
GPT-4.1-mini (Long Context) & 0.630 & 0.844 & 0.838 & 0.963 & 0.816 & 0.463 \\
Claude-Sonnet-4.5 (Long Context) & 0.588 & 0.868 & 0.838 & 0.981 & 0.752 & 0.598 \\
Gemini-3-Flash (Long Context) & 0.412 & 0.759 & 0.713 & 0.875 & 0.692 & 0.207 \\
GLM-4.6 (Long Context) & 0.670 & 0.886 & 0.843 & 0.947 & 0.804 & 0.415 \\
Qwen3.6-27b (Long Context) & 0.486 & 0.878 & 0.835 & 0.969 & \textbf{0.893} & 0.415 \\
DeepSeek-V4-Flash (Long Context) & 0.591 & 0.873 & 0.865 & 0.966 & 0.831 & 0.305 \\
\midrule
RAG (GPT-4.1-mini) (Turn-level) & 0.429 & 0.725 & 0.716 & 0.785 & 0.581 & 0.073 \\
RAG (GPT-4.1-mini) (Session-level) & 0.742 & 0.878 & 0.893 & 0.960 & 0.839 & 0.549\\
Mem0 & 0.226 & 0.715 & 0.632 & 0.938 & 0.382 & 0.085 \\
MemOS & 0.620 & 0.834 & 0.865 & 0.935 & 0.772 & 0.439 \\
Temp-LoRA & 0.052 & 0.236 & 0.190 & 0.343 & 0.060 & 0.012 \\
\bottomrule
\end{tabular}
}
\end{table*}

The previous section analyzes the results along the dimension of operation types. In this section, we shift to an orthogonal dimension by the evaluation probe types. Operation types characterize what kind of memory event occurred, such as remembering, forgetting, or updating. In contrast, probe types characterize which aspect of the memory capability is being evaluated, such as locating the triggering evidence, binding the correct target, determining the state transition, or applying the memory to a downstream task.
As illustrated in Table~\ref{tab:memoryops-v13-eval-type}, the adjacent setting again provides the strongest diagnostic performance. Qwen3.6-27B achieves the best adjacent scores on OperationTrace and StateTransition, while Claude-Sonnet-4.5 leads on TargetBinding, OperationApplication, StateTrajectory, and CandidateDisambiguation. These results show that Claude-Sonnet-4.5 and Qwen3.6-27B split dominance across probe types, with Qwen3.6-27B stronger at locating the triggering evidence and inferring the resulting state, and Claude-Sonnet-4.5 stronger at binding targets, applying memory to downstream tasks, tracking state trajectories, and rejecting distractors.

Across settings, CandidateDisambiguation is the easiest probe for strong models, with many adjacent and long-context methods scoring near ceiling, indicating that models can often reject an incorrect candidate when the alternatives are made explicit. In contrast, OperationTrace and StateTrajectory expose larger weaknesses.
Long-context scores on StateTrajectory remain far below target-binding and candidate-disambiguation scores for the same models. This contrast suggests that models can recognize relevant facts without reliably reconstructing the ordered memory-state trajectory that produced them.

Retrieval and memory baselines show the same separation between probe types. Session-level RAG substantially outperforms turn-level RAG on both OperationTrace and StateTrajectory. MemOS outperforms Mem0 across almost all probe types, with especially large gains on OperationTrace and StateTrajectory. Temp-LoRA remains weak across all probes, particularly on OperationTrace and StateTrajectory. These results support the central design of \model{} by demonstrating that final answer accuracy alone is insufficient to determine whether a system fails at event retrieval, target binding, state updating, or contextual use of the final memory.

\section{Conclusion and Future Work}

We introduce \model{}, a benchmark for evaluating long-term conversational
memory as a sequence of explicit lifecycle operations. Instead of treating
memory evaluation as final-answer question answering alone, \model{} represents
each memory-relevant event with a structured operation trace, evidence spans,
state transitions, and operation-targeted probes. This design makes it possible
to diagnose whether a system fails because it misses the triggering event,
binds the wrong target, applies an incorrect state transition, leaks forgotten
information, reuses stale values, or produces unsupported reflective memories.

Our experiments show that current systems remain far from uniformly reliable
under this operation-centric view. Strong adjacent-evidence models achieve high accuracy, but their strengths differ across operation detection,
provenance support, reflection precision, and operation families. Long-context
models retain substantial answer-level performance, yet show clear weaknesses
on state-trajectory reasoning, indicating that a larger context window does not guarantee faithful memory-state reconstruction. Retrieval and
persistent-memory baselines further reveal that performance depends on
how evidence is stored, retrieved, and interpreted: session-level retrieval is
substantially stronger than turn-level retrieval, MemOS outperforms Mem0 across
most diagnostic views, and parameter-update baselines remain unreliable for lifecycle memory control.

These findings suggest that future memory systems should maintain memory as an
auditable state rather than as an unstructured archive of past text. A useful
system should not only retrieve relevant user evidence, but also record which
operation it supports, what target it modifies, whether the value is active or
obsolete, and how later responses should respect that state. Future work can
extend \model{} along three directions. First, the benchmark can be scaled to
broader domains, languages, and longer multi-session histories while retaining
the same operation-level supervision. Second, the diagnostic metrics can be
combined with stronger human validation and judge-calibration studies, especially
for provenance and reflection quality. Third, \model{} can be used not only for
evaluation, but also as a development signal for memory controllers that
explicitly learn when to write, update, delete, retrieve, and justify memory
states. Together, these directions move long-term memory research toward systems
that are not merely accurate, but also controllable, inspectable, and aligned
with the intended lifecycle of user-provided information.

\bibliographystyle{unsrt}
\bibliography{reference}

\appendix

\end{document}